\documentclass[letterpaper]{article} 

\usepackage{aaai2026}
\nocopyright

\usepackage{times}  
\usepackage{helvet}  
\usepackage{courier}  
\usepackage[hyphens]{url}  
\usepackage{graphicx} 
\usepackage{natbib}  
\usepackage{caption} 
\usepackage{amsmath}
\usepackage{amssymb}
\usepackage{multirow}
\usepackage{algorithm}
\usepackage{algorithmic}
\usepackage{colortbl}      

\usepackage{booktabs}
\usepackage{float}
\usepackage{placeins}
\usepackage{newfloat}
\usepackage{listings}
\DeclareCaptionStyle{ruled}{labelfont=normalfont,labelsep=colon,strut=off} 
\floatstyle{ruled}
\newfloat{listing}{tb}{lst}{}
\floatname{listing}{Listing}
\title{GM-Skip: Metric-Guided Transformer Block Skipping for Efficient Vision-Language Models}
\author{
    \normalfont
    Lianming Huang\textsuperscript{1}\thanks{Equal contribution.} \quad
    Haibo Hu\textsuperscript{1}\footnotemark[1] \quad
    Qiao Li\textsuperscript{2} \quad
    Xin He\textsuperscript{3} \\
    Nan Guan\textsuperscript{1} \quad
    Chun Jason Xue\textsuperscript{2} \\
    \textsuperscript{1}City University of Hong Kong \quad
    \textsuperscript{2}MBZUAI \quad
    \textsuperscript{3}A*STAR \\
}
\usepackage{bibentry}

\begin{document}

\maketitle

\begin{abstract}
Transformer-based Vision-Language Models (VLMs) have achieved impressive performance on tasks such as image captioning, object recognition, and visual reasoning, but their high computational cost hinders deployment in latency-sensitive applications like autonomous driving. We introduce GM-Skip, a flexible and metric-adaptive framework for Transformer block skipping that accelerates VLM inference while preserving output quality. GM-Skip features a greedy, metric-guided block selection strategy that leverages metric feedback (e.g., accuracy, CIDEr) to identify redundant layers, along with a reverse-order deletion mechanism that preserves early foundational blocks to avoid performance collapse. To support diverse deployment needs, it incorporates a tunable trade-off between sparsity and performance via a score–sparsity balance objective. Experiments across multiple tasks and datasets, including COCO and CODA, demonstrate that GM-Skip consistently improves inference speed while maintaining task performance. On the COCO dataset, GM-Skip improves single-object classification accuracy on the \textit{Person} category from 19.1\% up to 87.3\% while skipping over 40\% of Transformer blocks. In real-world deployment, it achieves up to 45.4\% latency reduction on single-object detection when integrated into an autonomous vehicle running Autoware.Universe, validating the effectiveness of its skip configurations and confirming its practical value in accelerating real-world inference.

\end{abstract}


\section{Introduction}
\begin{figure}[t]
  \centering
\includegraphics[trim=0 50 400 30, clip,width=0.5\textwidth]{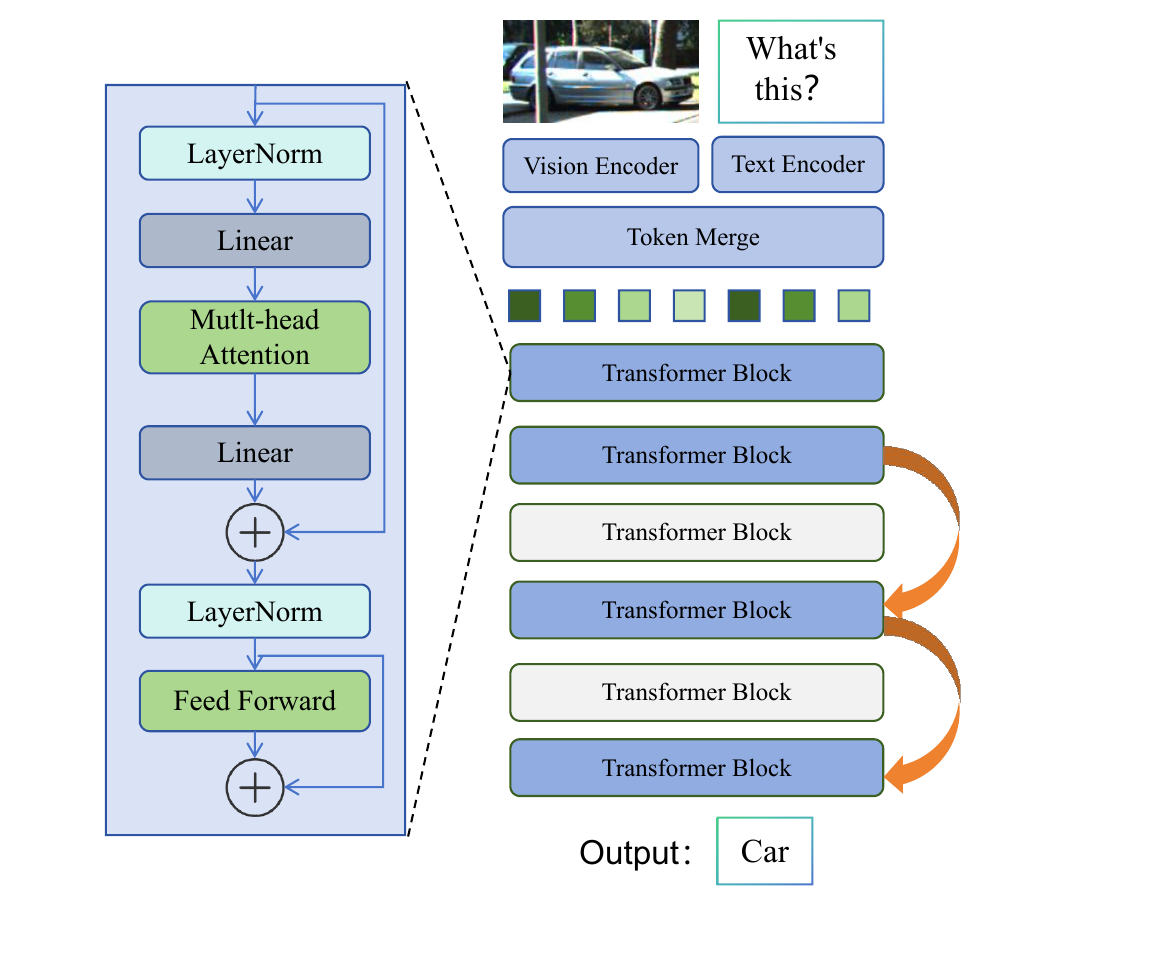}
\vspace{-20pt}
  \caption{Transformer block skipping where selected layers (gray ones) are skipped during inference to reduce computational cost while preserving output quality.}
  \label{pic:skip_sample}
\vspace{-10pt}
\end{figure}

Large language models (LLMs) have driven major advances in natural language understanding and generation~\cite{brown2020language}, evolving into increasingly multimodal systems that integrate vision and language~\cite{alayrac2022flamingo, gan2022vision}. VLMs now support tasks such as image captioning, Visual Question Answering (VAQ), and scene understanding, and have shown great potential in real-world applications like autonomous driving~\cite{zhou2024vision}. However, these models inherit a key limitation from LLMs: their large parameter scales impose high computational costs, making them unsuitable for real-time deployment in latency-sensitive scenarios~\cite{dhouib2025pact, gan2022vision, liao2024gpt}. To mitigate this, recent work explores skipping subsets of Transformer blocks during inference~\cite{shukor2024skipping, song2024sleb}, as illustrated in Figure~\ref{pic:skip_sample}. Compared to fine-grained weight pruning, which often results in unstructured sparsity and limited hardware gains~\cite{frantar2023sparsegpt}, block-level skipping exploits redundancy across adjacent layers~\cite{liu2023deja, din2023jump}, largely caused by residual connections~\cite{wang2020sparsert}. These methods reduce latency by skipping Transformer blocks that are less important to the task output, thus preserving accuracy.

Block-skipping techniques have seen limited adoption in VLMs.  Among existing block-skipping methods, Skip-MLLM~\cite{shukor2024skipping,shukor2024implicit} stands out as a representative and effective approach in the context of Vision-Language Models. However, its block selection relies on fixed interval pruning (e.g., removing every 2nd or 4th layer), which ignores the semantic contribution of each block. Such heuristics are task-agnostic and may discard functionally important layers, leading to unstable or degraded performance. As VLMs continue to expand into diverse applications, including captioning~\cite{alayrac2022flamingo, li2022blip}, object detection~\cite{li2022coda, dai2023instructblip}, and visual reasoning~\cite{chen2023vlm}, the demand for an effective and adaptive skipping architecture becomes more pressing. Instead of applying LLM-derived logic directly, we revisit the problem from a VLM-centric perspective, and design a more flexible selection mechanism that accounts for task structure, metric diversity, and representation dynamics.

In addition, most skipping strategies are grounded in the assumption that similar intermediate representations imply redundant computation~\cite{shukor2024skipping,shukor2024implicit}. Yet they rarely consider how to select among multiple redundant candidates. Drawing inspiration from early-exit literature~\cite{schuster2022confident, huang2024raee}, we hypothesize that earlier blocks play foundational roles in grounding vision-language alignment. Our experiments show that removing early blocks can lead to catastrophic performance collapse, while skipping later blocks in reverse order yields more stable outputs. In our preliminary results, we observe that forward deletion causes accuracy to drop sharply—down to near zero—when removing the fourth layer, whereas reverse-order deletion maintains stable performance throughout. 
This suggests that, beyond redundancy, block position is a key factor in maintaining performance, and motivates a depth-aware deletion strategy.

While LLM sparsity techniques are typically benchmarked on standard datasets, where evaluation metrics closely align with their deployment goals~\cite{achiam2023gpt4v,dubey2024llama}, VLMs differ in that their downstream applications often involve real-world visual perception tasks~\cite{zhou2024vision,alayrac2022flamingo,li2022coda}. Therefore, the application of skip mechanisms in VLMs requires validation in real-world scenarios to ensure their effectiveness beyond offline benchmarks~\cite{zhou2024vision,jiang2024effectiveness, zhang2025mole}.

To address the aforementioned challenges, we propose GM-Skip, a novel block skipping framework tailored to the diverse tasks of VLMs. GM-Skip introduces a flexible, metric-based block selection strategy, where skip decisions are made using a greedy algorithm. At each step, we skip the Transformer block whose removal results in an improvement or a minimal decrease in the task-specific metric. To enhance stability, GM-Skip adopts a reverse-order deletion strategy, which delays inference collapse by retaining the essential early layers. Furthermore, the score–sparsity balance mechanism provides flexible control between accuracy and efficiency: either maintaining performance comparable to the full-layer model, or accepting moderate score degradation in exchange for higher sparsity and faster inference. Motivated by the growing integration of VLMs into autonomous driving systems, as demonstrated by recent efforts from companies like Xiaomi~\cite{coreteam2025mimovltechnicalreport} and Li Auto~\cite{tian2024drivevlm}, we evaluate GM-Skip in real-world autonomous driving scenarios to validate its practical effectiveness.
Our contributions are as follows: 
\begin{itemize}
    \item We present GM‑Skip, a flexible block-skipping framework for VLMs that supports task-specific evaluation metrics and adapts to multiple vision-language task types, such as image classification and caption generation.
    \item We propose a reverse-order deletion strategy that reduces the risk of output collapse when skipping blocks in VLMs, as it avoids removing critical early layers. In addition, our tunable score-to-sparsity balance mechanism enables flexible tradeoffs between computational efficiency and performance preservation.
    \item We demonstrate the practical effectiveness of our framework through deployment on a real autonomous vehicle, where GM-Skip achieves up to 45.4\% inference latency reduction in single-object detection under the High Sparsity setting.
\end{itemize}
    
\section{Related Work}
\subsection{Visual-Language Model}
The rapid progress of LLMs~\cite{touvron2023llama, chiang2023vicuna, yang2024qwen2} has laid a strong foundation for the development of large-scale VLMs~\cite{alayrac2022flamingo, bai2023qwenvl, chen2023shikra, achiam2023gpt4v, reid2024gemini}. Among them, models like Flamingo~\cite{alayrac2022flamingo} pioneered the integration of frozen LLMs with image encoders for few shot multimodal learning. Closed-source solutions like GPT-4V~\cite{achiam2023gpt4v} and Gemini-1.5~\cite{reid2024gemini} further demonstrated remarkable multimodal capabilities through large-scale proprietary training. In contrast, open-source models such as the LLaVA series~\cite{liu2024llava, liu2024llavanext, xu2024llavauhd} inject image features into LLMs via visual projectors, but at the cost of excessive visual tokens, limiting efficiency. Techniques like InternVL~\cite{chen2024internvl}, MiniCPM-V~\cite{yao2024minicpm}, and BRAVE~\cite{kar2024brave} address this by compressing or dynamically segmenting visual information, often via cross attention or multi-encoder schemes. Additionally, instruction tuning~\cite{dai2023instructblip, liu2024llava} enables these models to better align with user prompts and perform natural conversations.

\subsection{Sparsity and Efficient Inference in Large Models}
Improving the inference efficiency of large models has been extensively studied through pruning and sparsity techniques. Unstructured pruning removes individual weights~\cite{frantar2023sparsegpt, sun2023wanda}, while structured pruning, such as 2:4 sparsity~\cite{frantar2022optimal}, aligns better with hardware but often yields limited real-world speedup. Recent efforts move toward skipping entire Transformer blocks based on the redundancy of intermediate representations~\cite{liu2023deja}. SLEB~\cite{song2024sleb} introduces a token-level feedback-driven method to eliminate redundant blocks in LLMs without retraining, achieving significant efficiency gains. In the multimodal setting, Shukor et al.~\cite{shukor2024implicit} apply similar skipping strategies to VLMs, targeting both FFN and attention modules. However, these methods largely inherit LLM-based assumptions and fail to fully leverage the task diversity and metric heterogeneity intrinsic to VLMs. As a result, skip based Transformer compression in VLMs remains underdeveloped and lacks tight integration with the unique characteristics of multimodal tasks.
\section{Motivation}
\begin{figure}[h]
\centering
\includegraphics[width=1.0\linewidth]{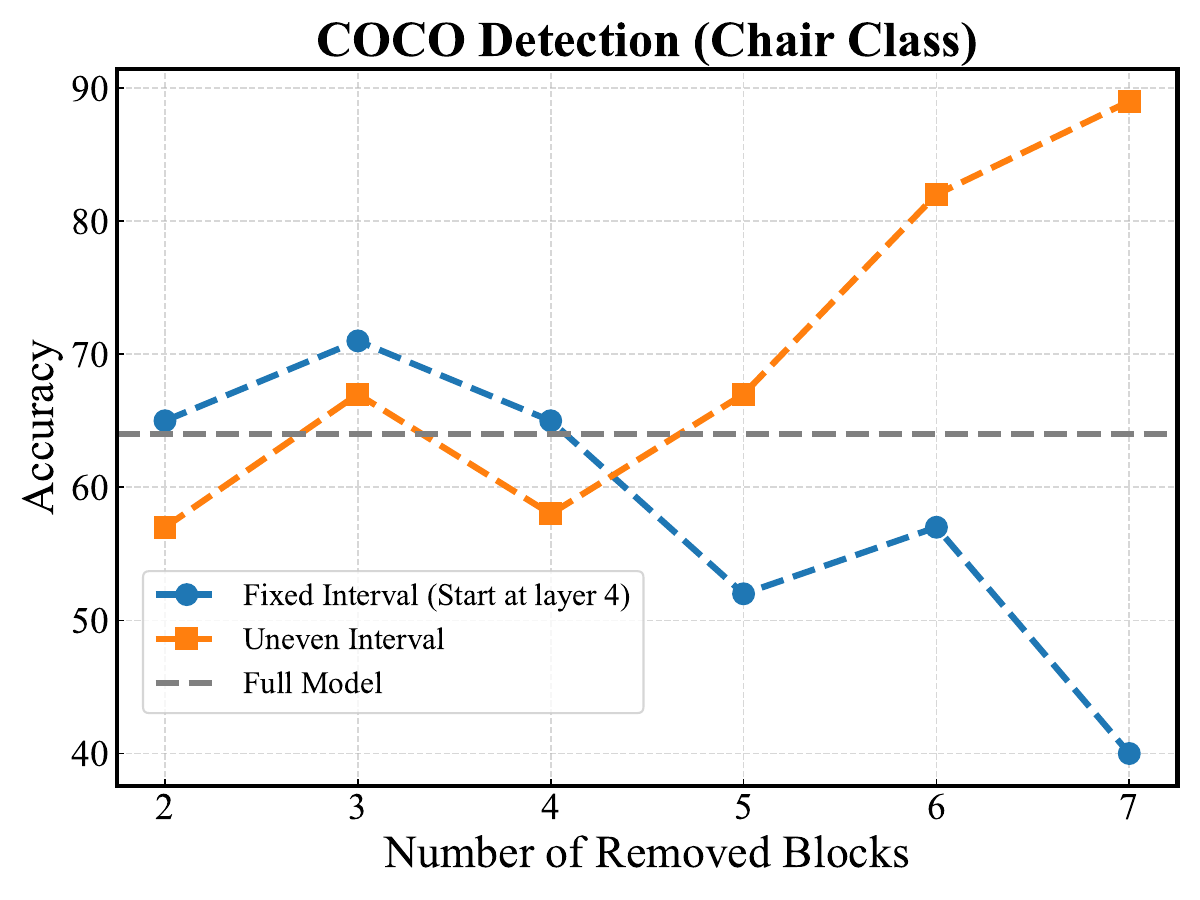}
\vspace{-15pt}
\caption{Accuracy comparison between uneven interval block removal and fixed interval removal on the COCO chair class. }
\label{fig:motivation_1}
\vspace{-10pt}
\end{figure}

Transformer block skipping relies on the observation that deeper layers often yield semantically redundant representations~\cite{song2024sleb}. In the VLM domain, Skip-MLLM~\cite{shukor2024implicit} adopts a fixed-interval deletion strategy, removing every 2nd, 4th, or 8th layer without feedback. As illustrated in Figure~\ref{fig:motivation_1}, this heuristic strategy (Fixed Interval) achieves limited accuracy improvements and performs comparably to randomly sampled deletions (Uneven Interval), indicating that naive block selection fails to reflect true utility. To address this, we propose a metric-flexible framework that evaluates block utility via its direct impact on task-specific performance. This enables a \textit{pluggable feedback interface} that integrates diverse evaluation metrics—such as accuracy, CIDEr, and set-level precision—allowing greedy block selection to align better with downstream objectives.

\begin{figure}[t]
\centering
\includegraphics[width=1.0\linewidth]{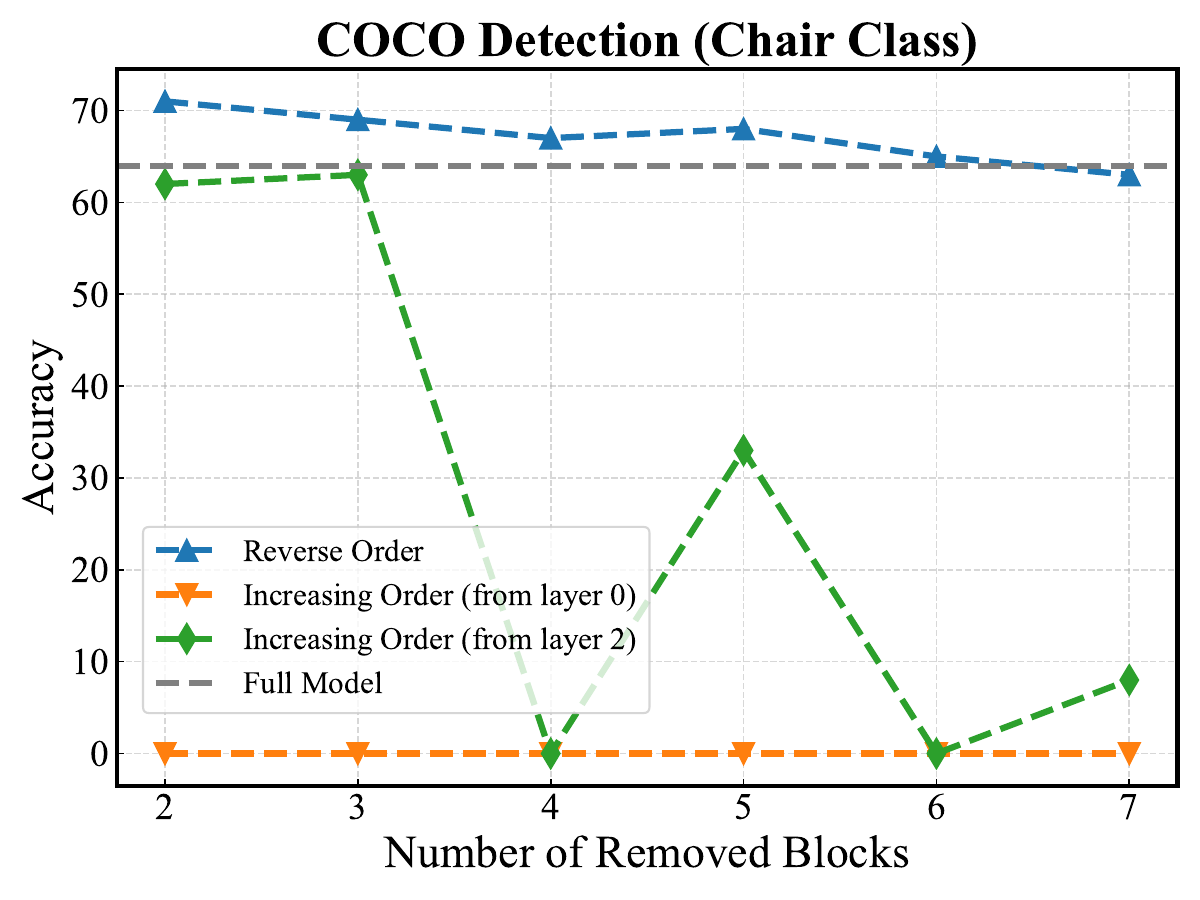}
\vspace{-15pt}
\caption{Accuracy comparison of different block removal strategies on the COCO chair class, showing that reverse order deletion maintains performance better than increasing order from early layers. 
}
\label{fig:motivation_2}
\vspace{-10pt}
\end{figure}

Furthermore, existing methods often assume that Transformer blocks are order-invariant, overlooking their functional asymmetry. Our controlled experiments reveal that early layers are crucial for low-level visual grounding, while deeper layers contribute incrementally. As shown in Figure~\ref{fig:motivation_2}, reverse-order deletion (i.e., removing deeper layers first) yields robust and stable performance. In contrast, deleting from shallow layers (e.g., increasing order from layer 0 or 2) leads to catastrophic collapse in accuracy, even with only a few blocks removed. These results underscore the need for an order-aware skipping strategy that protects structurally important blocks. Finally, we observe a non-linear performance drop when too many blocks are skipped, highlighting the importance of a tunable trade-off mechanism between sparsity and accuracy. These insights collectively motivate our design of GM-Skip: a metric-driven skipping framework tailored for VLMs.

\begin{figure}[t]
  \centering
\includegraphics[width=1.0\linewidth]{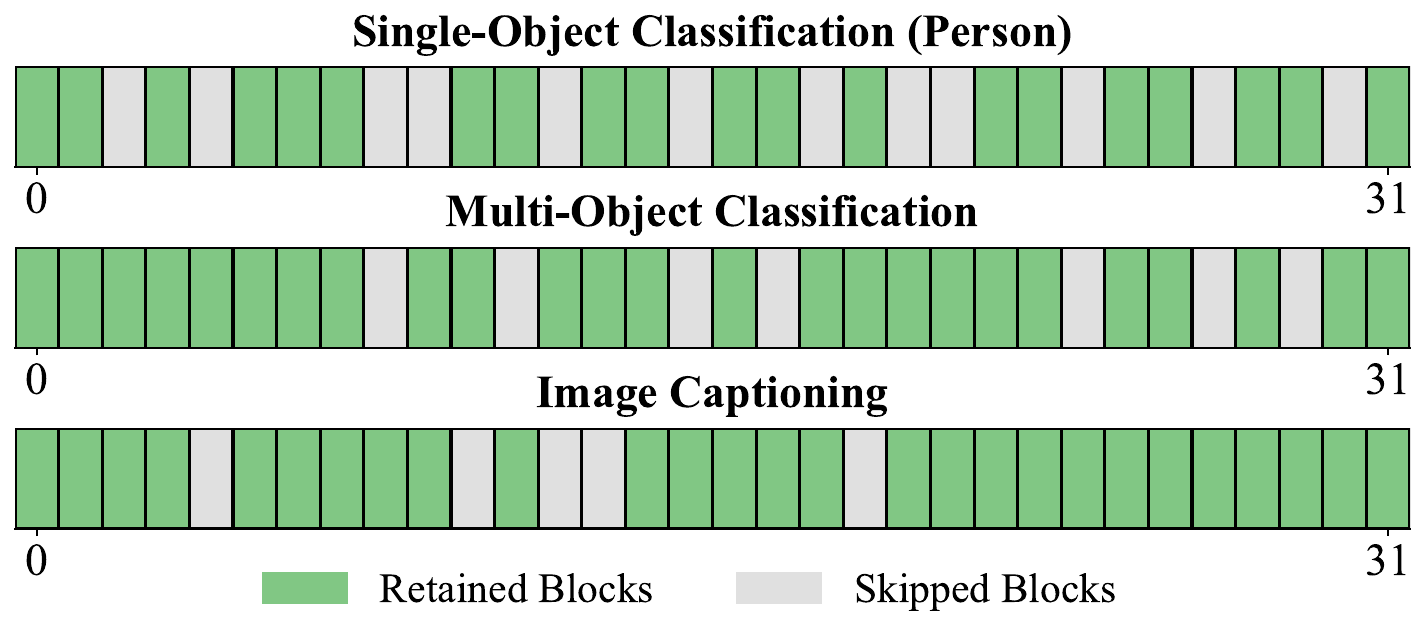}
\vspace{-20pt}
  \caption{Visualization of selected Transformer blocks across three VLM tasks—simpler tasks like classification allow more aggressive skipping, while generation tasks retain more blocks due to their complexity.}
  \label{fig:selected_blocks}
\end{figure}

Beyond algorithmic design, it is also important to consider how task complexity influences the effectiveness of block skipping. As illustrated in Figure~\ref{fig:selected_blocks}, our preliminary experiments show that block skipping is more effective in tasks involving fewer generated tokens, where models exhibit clearer redundancy and higher tolerance to sparsity. Motivated by this observation, we select three representative VLM tasks of increasing difficulty, Single-Object Classification, Multi-Object Classification, and Image Captioning, to evaluate our method under varying levels of semantic complexity. The first task provides a strong binary supervision signal that enables precise attribution of block importance. The second introduces partial correctness and multi-entity grounding, challenging the skip strategy to retain accuracy across distributed object regions. The third imposes the greatest difficulty, as coherent caption generation depends heavily on deep-layer semantic refinement. These tasks together allow us to systematically examine how block skipping behaves across diverse VLM objectives. 
\begin{figure*}[!t]
\centering
\vspace{-80pt}
\includegraphics[width=1\textwidth, height=0.4\textheight]{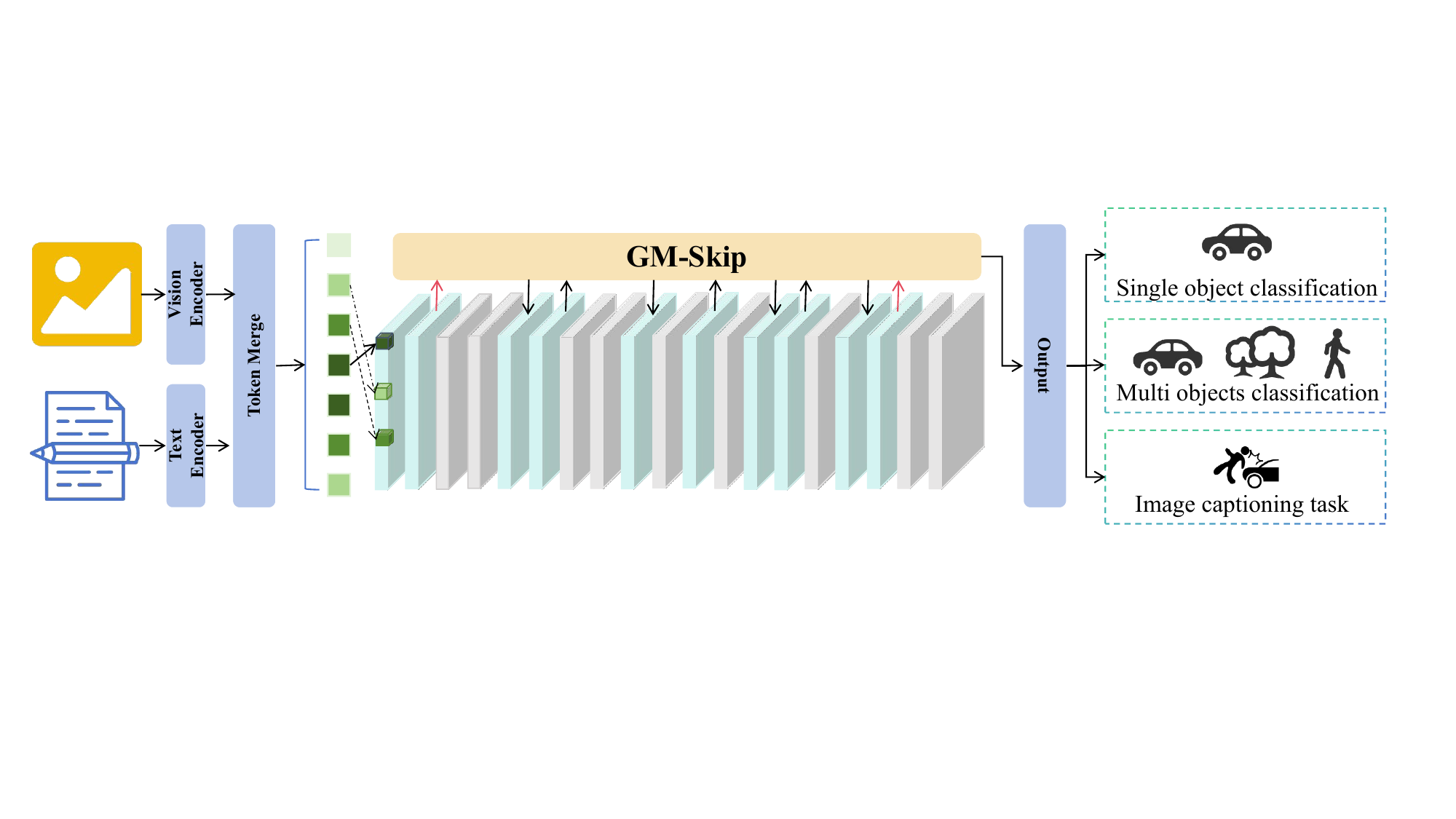}
\vspace{-100pt}
\caption{Overview of our skipping framework. Given an image and text prompt, visual and textual features are fused into a unified token sequence. A Skipping Control module dynamically selects Transformer blocks based on task type and token behavior, enabling efficient inference across classification and generation tasks.}
\label{fig:overview}
\end{figure*}
\section{Methodology}
This section presents the proposed GM-Skip framework for efficient and task-adaptive Transformer block skipping in VLMs, with Figure~\ref{fig:overview} illustrating the overall architecture. 
 
\subsection{Problem Formulation}
Denote $(x_{\text{img}}, x_{\text{text}})$ as the paired image-text input of a Vision-Language Model (VLM) $f$, which is encoded into joint tokens $X = \text{Enc}(x_{\text{img}}, x_{\text{text}})$. These tokens are processed by a sequence of $L$ Transformer blocks $\{B_1, B_2, \dots, B_L\}$ to produce the final output $\hat{y}$.

To accelerate inference, we aim to skip a subset of blocks while maintaining task performance. Let $\mathcal{S} \subseteq \{1, \dots, L\}$ denote the set of retained (executed) blocks, and $f_{\mathcal{S}}(X)$ the resulting output. Given a calibration dataset $\mathcal{D}$ and a task-specific evaluation metric $\mathcal{M}(f_{\mathcal{S}}, \mathcal{D}) \in \mathbb{R}$ (e.g., CIDEr or accuracy), our objective is to find the optimal block configuration $\mathcal{S}^*$ by minimizing:

\begin{equation}
\mathcal{S}^* = \arg\min_{\mathcal{S} \subseteq \{1, \dots, L\}} 
\left(\mathcal{M}(f_{\mathcal{L}}, \mathcal{D})-\mathcal{M}(f_{\mathcal{S}}, \mathcal{D}) \right)
\label{eq:skip-objective}
\end{equation}

Here, $\mathcal{M}$ is computed over $\mathcal{D}$ to approximate downstream task performance. This formulation guides the selection of $\mathcal{S}^*$ to retain accuracy.

\subsection{Task Metrics}
\paragraph{(1) Single-Object Classification.}  
We select single-object classification as a representative of low-complexity VLM tasks. This task is not only fundamental but also widely studied in recent research that integrates VLMs with smaller models for auxiliary perception or decision making. In this setting, the model predicts a single object label $\hat{y} \in \mathcal{C}$ from an image-text input, and we use top-1 accuracy as the evaluation metric:

\begin{equation}
\mathcal{M}_{\text{cls}}(\hat{y}, y) = \delta(\hat{y}, y),
\label{eq:acc}
\end{equation}

where $\delta(\hat{y}, y) = 1$ if $\hat{y} = y$, and $0$ otherwise.

Unlike complex metrics such as CIDEr or BLEU, which require heuristic matching or partial credit scoring, this metric offers direct and trustworthy feedback on the model's behavior. As a result, it enables clear observation of how effectively our block-skipping strategy maintains performance while adhering to strict correctness constraints, making it an ideal benchmark for evaluating skip decision efficacy.

\paragraph{(2) Multi-Object Set Classification.}  
We select multi-object classification to represent medium complexity VLM tasks. These tasks require the model to identify multiple relevant entities in a scene, forming a prediction set $\hat{Y} \subseteq \mathcal{C}$. The evaluation emphasizes not only correctness but also partial completeness, making it more tolerant than strict single-label classification. We use set level precision as the metric:
\begin{equation}
\mathcal{M}_{\text{set}}(\hat{Y}, Y) = \frac{|\hat{Y} \cap Y|}{|Y|}, \quad \text{where } Y = \{y_1, y_2, \dots, y_n\}
\label{eq:set_acc}
\end{equation}

where $\hat{Y}, Y \subseteq \mathcal{C}$ denote the sets of predicted and ground-truth labels. Although it weakens the binary supervision signal available in single-object tasks, this setup introduces a stronger challenge for skip strategies, as the model must retain discriminative capacity across diverse object categories under reduced computational depth.

\paragraph{(3) Image Captioning.}  
We select image captioning as a representative high-complexity task, where the VLM must generate a natural language description $\hat{c}$ for a given image. This task requires fine-grained visual-textual alignment and abstract scene understanding. For evaluation, we adopt the widely used CIDEr score~\cite{vedantam2015cider}, defined as:

\begin{equation}
\mathcal{M}_{\text{cap}}(\hat{c}, \mathcal{C}) = \text{CIDEr}(\hat{c}, \mathcal{C}),
\label{eq:cider}
\end{equation}

which measures the n-gram similarity between the generated caption $\hat{c}$ and a set of human written references $\mathcal{C}$, with TF-IDF weighting to emphasize salient content.

Unlike classification tasks with binary or set-based correctness criteria, captioning metrics rely on heuristic matching and semantic overlap, introducing ambiguity in the supervision signal.
\begin{algorithm}[t]
\caption{Greedy Skipping Algorithm with Reverse and Balance}
\label{alg:greedy}
\begin{algorithmic}[1]
\REQUIRE Pretrained VLM $f$, calibration dataset $\mathcal{D}$, total blocks $L$, max removal $K$, metric function $\mathcal{M}$, trade off weight $\lambda$
\STATE Initialize: $\mathcal{S} \leftarrow \{1, 2, \dots, L\}$ \hfill \textit{// Full set of blocks}
\FOR{$i = 1$ to $K$}
    \FOR{each $b \in \mathcal{S}$}
        \STATE $\mathcal{S}_b \leftarrow \mathcal{S} \setminus \{b\}$
        \STATE Compute $\Delta_b \leftarrow \mathcal{M}(f_{\mathcal{S}}, \mathcal{D}) - \mathcal{M}(f_{\mathcal{S}_b}, \mathcal{D})$
    \ENDFOR
    \STATE $b^* \leftarrow \arg\min_b \Delta_b$ \hfill
    \textit{// Least total loss}
    \STATE Tie-breaker: choose largest $b^*$ if multiple blocks share the same $\Delta_b$
\IF{$\mathcal{M}(f_{\mathcal{S}_{b^*}}, \mathcal{D}) \geq  \lambda\cdot \mathcal{M}(f_{\mathcal{L}}, \mathcal{D})$}
    \STATE $\mathcal{S} \leftarrow \mathcal{S} \setminus \{b^*\}$
\ELSE
    \STATE \textbf{break}
\ENDIF
\ENDFOR
\RETURN Skipped block set $\{1, \dots, L\} \setminus \mathcal{S}$
\end{algorithmic}
\end{algorithm}

\subsection{Metric-Guided Greedy Block Skipping}
To efficiently identify the most dispensable Transformer blocks, we employ a greedy, metric-guided block selection strategy. At each iteration, we evaluate the marginal impact of removing each retained block by computing the change—either decrease or increase—in task-specific performance (e.g., accuracy or CIDEr). The key intuition is that, due to the residual accumulation, Transformer layers contribute incrementally and unequally to the final output. Rather than estimating importance from static heuristics, we dynamically assess utility under the current context of retained layers. This greedy approach works by locally optimizing the block removal decision at each step: for the current set of active blocks, we simulate the exclusion of each candidate and observe its effect on model performance. Since the remaining blocks define a new intermediate representation, the evaluation is state-aware and adapts to each removal. Although this strategy does not guarantee a globally optimal skip configuration, it effectively approximates an optimal solution by prioritizing low-impact deletions under updated conditions.

Formally, given a pretrained VLM composed of $L$ Transformer blocks, we define a retained block set $\mathcal{S} \subseteq \{1, \dots, L\}$. For each candidate block $b \in \mathcal{S}$, we assess its significance by computing the drop or increase in a task-specific metric $\mathcal{M}$ on a calibration set $\mathcal{D}$ when the block is removed:

\begin{equation}
\Delta_b = \mathcal{M}(f_{\mathcal{S}}, \mathcal{D}) - \mathcal{M}(f_{\mathcal{S} \setminus \{b\}}, \mathcal{D}).
\label{eq:greedy_delta}
\end{equation}
This \textit{delta-based evaluation} enables the algorithm to directly measure functional degradation relative to real task metrics: classification accuracy, CIDEr, or set precision. By doing so, our approach eliminates the need for metric-specific threshold tuning or scale normalization. At each iteration, the algorithm greedily removes the block with the smallest $\Delta_b$, i.e., the one whose removal causes the least performance degradation.

\subsection{Reverse Deletion and Performance-Sparsity Balance}
When multiple blocks yield comparable performance impacts during greedy selection, we apply a reverse-order preference by selecting the block with the highest layer index. This ensures that deeper blocks are considered for removal before earlier ones, preserving low-level feature extractors in the shallower layers.

To further guide block selection under ambiguous cases, we incorporate a sparsity-aware balancing mechanism that considers both the performance drop and the overall reduction in block count. Together, these two strategies form a unified decision rule: among all candidates with minimal task impact, we prefer those that (1) appear later in the model and (2) contribute more to sparsity. This cooperative design allows each greedy iteration to choose a block that is both structurally safe to remove and incrementally improves efficiency.

We incorporate a tunable balance between accuracy and sparsity by modifying the objective in Eq.~\eqref{eq:balance} into a weighted formulation:

\begin{equation}
\mathcal{M}(f_{\mathcal{S}_{b^*}}, \mathcal{D}) \geq \lambda \cdot \mathcal{M}(f_{\mathcal{L}}, \mathcal{D})
\label{eq:balance}
\end{equation}

Here, $\mathcal{M}(\cdot, \mathcal{D})$ is a task-specific evaluation metric (e.g., accuracy or CIDEr) computed on the calibration set $\mathcal{D}$, and $f_{\mathcal{L}}$ denotes the original model with all blocks retained. The hyperparameter $\lambda \in [0,1]$ specifies the minimum acceptable performance ratio relative to the full model. For example, setting $\lambda = 0.9$ enforces that each pruning decision must retain at least 90\% of the original metric value. This formulation offers a simple and interpretable performance-based constraint that ensures model robustness during block skipping.

\begin{table*}[!htbp]
\centering
\caption{Comparison of block skipping strategies on Single-object classification (part 1: Person to Cup).}
\label{table:single_object_1}
\renewcommand{\arraystretch}{1}
\resizebox{\textwidth}{!}{%
\begin{tabular}{c|ccc|ccc|ccc|ccc}
    \hline
    \multirow{2}{*}{Methods} 
    & \multicolumn{3}{c|}{Person} 
    & \multicolumn{3}{c|}{Car} 
    & \multicolumn{3}{c|}{Chair} 
    & \multicolumn{3}{c}{Cup} \\
    \cline{2-13}
    & Acc. & Sps. & Lat. 
    & Acc. & Sps. & Lat. 
    & Acc. & Sps. & Lat. 
    & Acc. & Sps. & Lat. \\
    \hline
    Baseline(Full-layer)
    & 19.10 & 0.00 & 0.2038 
    & 60.70 & 0.00 & 0.1825 
    & 62.50 & 0.00 & 0.1624 
    & 11.60 & 0.00 & 0.1547 \\
    \hline
    Skip-MLLM 
    & 0.00  & 31.25 & 0.1071 
    & 53.40 & 31.25 & 0.1333 
    & 30.00 & 31.25 & 0.1210 
    & 5.20  & 31.25 & 0.1118 \\
    \hline
    GM-Skip (High Perf.) 
    &87.30 & 40.63 & 0.1062 
    & 95.20 & 34.38 & 0.1277 
    &98.40 & 40.63 & 0.1064 
    & 68.70 & 37.50 & 0.1066 \\
    \hline
    GM-Skip (High Sps.) 
    & 37.30 & 56.25 & 0.0771 
    & 67.00 & 59.38 & 0.0889 
    & 88.20 & 62.50 & 0.0758 
    & 31.00 & 78.13 & 0.0528 \\
    \hline
\end{tabular}
}
\end{table*}

\begin{table*}[!htbp]
\centering
\caption{Comparison of block skipping strategies on Single-object classification (part 2: Bird to Cow).}
\label{table:single_object_2}
\renewcommand{\arraystretch}{1}
\resizebox{\textwidth}{!}{%
\begin{tabular}{c|ccc|ccc|ccc|ccc}
    \hline
    \multirow{2}{*}{Methods} 
    & \multicolumn{3}{c|}{Bird} 
    & \multicolumn{3}{c|}{Boat} 
    & \multicolumn{3}{c|}{Sheep} 
    & \multicolumn{3}{c}{Cow} \\
    \cline{2-13}
    & Acc. & Sps. & Lat. 
    & Acc. & Sps. & Lat. 
    & Acc. & Sps. & Lat. 
    & Acc. & Sps. & Lat. \\
    \hline
    Baseline(Full-layer)
    & 39.40 & 0.00 & 0.1721 
    & 1.20  & 0.00 & 0.1768 
    & 0.70  & 0.00 & 0.1745 
    & 56.00 & 0.00 & 0.1776 \\
    \hline
    Skip-MLLM 
    & 7.30  & 31.25 & 0.1252 
    & 1.90  & 31.25 & 0.1279 
    & 0.60  & 31.25 & 0.1273 
    & 0.20  & 31.25 & 0.1285 \\
    \hline
    GM-Skip (High Perf.) 
    & 94.30 & 43.75 & 0.1079 
    & 50.50 & 46.88 & 0.1058 
    & 51.30 & 34.38 & 0.1225 
    & 69.10 & 34.38 & 0.1020 \\
    \hline
    GM-Skip (High Sps.) 
    & 50.60 & 53.13 & 0.0936 
    & 26.60 & 87.50 & 0.0436 
    & 15.80 & 71.88 & 0.0663 
    & 59.70 & 40.63 & 0.1157 \\
    \hline
\end{tabular}
}
\end{table*}
\subsection{Deployment in Autonomous Driving}
To enable real-world deployment, we integrate our skipping-aware VLM framework into a physical autonomous vehicle running Autoware.Universe stack. Images are captured from the front-facing camera via ROS2 at 10Hz and subsampled to 1Hz for real-time inference. We first train the block skipping configuration on the CODA dataset, which provides diverse and challenging driving scenes with visual-linguistic annotations. The learned skip strategy is then directly applied to the onboard system, allowing comparative evaluation between full-layer and skip-layer modes under real driving conditions.

To evaluate task performance, we implement three representative VLM tasks onboard:
\begin{itemize}
    \item \textbf{Single-Object Classification}: For common road agents such as \textit{persons} and \textit{cars}, we conduct object-level classification using VLMs by directly querying the image with relevant prompts, and compare the model outputs under both skip and non-skip configurations.
    \item \textbf{Multi-Object Classification}: We prompt the VLM with the entire image to perform set-level classification, enabling evaluation of model performance in complex spatially entangled scenes.
    \item \textbf{Image Captioning}: For captioning, we directly input the full subsampled frame and assess the generated captions via human inspection and qualitative analysis. Since the CODA dataset does not contain ground-truth captions, we directly test using the skip configuration trained on the COCO dataset.
\end{itemize}
\section{Experiment}
We conduct comprehensive experiments to evaluate the effectiveness of our proposed skipping strategy across both general purpose and domain specific VLMs. Our goals are twofold: (1) to verify that our method can reduce inference latency while preserving output quality across different task types, and (2) to demonstrate the real-world applicability of our approach in autonomous driving scenarios. To assess task specific block selection in autonomous driving, we further train skipping configurations on the CODA dataset and deploy them in a real-world vehicle equipped with Autoware.Universe. Please refer to the Technical Appendix for additional ablation results and qualitative visualizations and further details of the experimental results.

\subsection{Implementation Details}
\paragraph{Datasets.}  
We use two datasets in our experiments. For general-purpose VLM evaluation, we adopt Microsoft COCO~\cite{lin2014microsoft}, a large-scale benchmark with over 120,000 images annotated for both classification and captioning, suitable for analyzing output length effects. For domain-specific deployment, we use CODA~\cite{coda123}, a real-world autonomous driving dataset focused on safety-critical corner cases with rich annotations for object detection under challenging conditions.

\paragraph{Experiment Setup.}  
All experiments are conducted on NVIDIA A100 GPUs with 80GB memory. For real-world autonomous driving evaluations, we deploy the model on an industrial-grade onboard computer equipped with an NVIDIA RTX 3090 GPU. 
    
\paragraph{VLM Backbone.}  
We adopt LLaVA~\cite{li2024llava}, a widely used open-source Vision-Language Model that combines a CLIP-based visual encoder with a Vicuna language decoder. We choose LLaVA due to its open accessibility, extensive adoption in the community, and modular architecture that facilitates transformer block manipulation during inference.

\subsection{Result}
\paragraph{Single-object classification.} 
We evaluate block skipping strategies across eight COCO categories using accuracy, sparsity, and latency as metrics. GM-Skip (High Perf.) consistently achieves high accuracy (e.g., \textbf{87.3\%} on Person, \textbf{98.4\%} on Chair, and \textbf{94.3\%} on Bird) with moderate sparsity\textbf{ (34\%--47\%)}, while GM-Skip (High Sps.) yields higher sparsity levels\textbf{ (up to 87.5\%)} with reduced latency (as low as \textbf{0.0436s} on Boat). In contrast, Skip-MLLM maintains lower accuracy under similar sparsity budgets (e.g., \textbf{0\%} on Person, \textbf{7.3\%} on Bird). The baseline without skipping has full accuracy but incurs the highest latency.

\paragraph{Multi-object classification.}We evaluate GM-Skip on the multi-object classification task to assess its effectiveness under more complex, set-based outputs. As shown in Table~\ref{tab:multi_object_comparison}, GM-Skip outperforms both the baseline and existing block skipping methods across multiple trade-off settings. In a high-sparsity configuration, GM-Skip achieves a sparsity of \textbf{25.00\%} with a competitive accuracy of \textbf{60.09}, significantly surpassing Skip-MLLM (Int=4), which suffers a drastic accuracy drop (\textbf{48.42}) despite similar or higher sparsity levels. In the high-performance regime, GM-Skip attains the highest accuracy of \textbf{66.01}, even outperforming the non-sparse baseline (\textbf{60.27}), while still reducing latency from \textbf{1.0218s} to \textbf{0.7781s}. 
\begin{table*}[!htbp]
\centering
\caption{Comparison of block skipping strategies on Real-World. (Ski. = Skipped block index)}
\label{table:real_world}
\renewcommand{\arraystretch}{1}
\resizebox{\textwidth}{!}{%
\begin{tabular}{l|ccp{4cm}|ccp{4cm}|ccp{3cm}}
    \hline
    \multirow{2}{*}{Method} 
    & \multicolumn{3}{c|}{Detection - Single} 
    & \multicolumn{3}{c|}{Detection - Multi} 
    & \multicolumn{3}{c}{Captioning} \\
    \cline{2-10}
    & Sps. & Lat. & Ski. 
    & Sps. & Lat. & Ski. 
    & Sps. & Lat. & Ski. \\
    \hline
    Full-layer 
    & 0.00 & 0.6006 & [] 
    & 0.00 & 1.4265 & [] 
    & 0.00 & 1.1293 & [] \\
    \hline
    High Perf 
    & 34.38 & 0.4140 & [29, 16, 27, 21, 28, 24, 20, 17, 2, 6, 7]
    & 28.13 & 1.0589 & [24, 28, 23, 13, 25, 6, 21, 17, 9] 
    & 9.38 & 0.8683 & [13, 19, 12] \\
    \hline
    High Sps 
    & 50.00 & 0.3280 & [29, 16, 27, 21, 28, 24, 20, 17, 2, 6, 7, 25, 8, 11, 12, 26] 
    & 31.25 & 1.0170 & [24, 28, 23, 13, 25, 6, 21, 17, 9, 18] 
    & 21.88 & 0.7106 & [13, 19, 12, 4, 10, 23, 21]\\
    \hline
\end{tabular}
}
\end{table*}

\begin{table}[t]
\centering
\caption{Comparison of block skipping strategies on Multi-object classification. Lat. = Latency (s), Sps. = Sparsity (\%), Int. = Interval, Accs = Accuracy set(\%).}
\label{tab:multi_object_comparison}
\small
\setlength{\tabcolsep}{6pt}
\begin{tabular}{lcccc}
\toprule
Method & Lat. & \#Param & Sps. & Accs↑ \\
\midrule
Skip-MLLM (Int.=4) & 0.8114 & 6.7B & 21.88 & 48.42 \\
Skip-MLLM (Int.=7) & 0.8848 & 6.7B & 12.50 & 51.06 \\
GM-Skip (High Sps.) & 0.6881 & 6.7B & \textbf{25.00} & 60.09 \\
Baseline & 1.0218 & 6.7B & 0.00 & 60.27 \\
GM-Skip (High Perf.) & 0.7781 & 6.7B & 12.50 & \textbf{66.01} \\
\bottomrule
\end{tabular}
\end{table}

\begin{table}[t]
\centering
\caption{Comparison of block skipping strategies on COCO captioning.}
\label{tab:caption}
\small
\setlength{\tabcolsep}{3pt}
\begin{tabular}{lcccc}
\toprule
Method & Lat.  & \#Param & Sps. & CIDEr (test) ↑ \\
\midrule
Skip-MLLM(Int.=6) & 0.5087 & 6.7B & 15.63 & 97.09 \\
GM-Skip (High Sps.) & 0.4413 & 6.7B & \textbf{21.88} & 97.22 \\
Baseline & 0.6546 & 6.7B & 0.00 & 98.43 \\
Skip-MLLM(Int.=10) & 0.5490 & 6.7B & 9.38 & 100.84 \\
GM-Skip (High Perf.) & 0.5029 & 6.7B & 9.38 & \textbf{111.07} \\
\bottomrule
\end{tabular}
\end{table}

\paragraph{Image Captioning.}
We evaluate GM-Skip on the COCO image captioning benchmark using CIDEr as the primary evaluation metric. As shown in Table~\ref{tab:caption}, GM-Skip achieves a CIDEr score of \textbf{111.07} with \textbf{9.38\%} sparsity, outperforming both the baseline (\textbf{98.43} with 0\% sparsity) and Skip-MLLM (\textbf{100.84} at 9.38\% sparsity with fixed interval = 10). Under a high-sparsity setting, GM-Skip maintains strong performance (\textbf{97.22} at \textbf{21.88\%} sparsity), while Skip-MLLM (Int=6) drops to \textbf{97.09} with significantly lower sparsity (\textbf{15.63\%}). 
\subsection{Experiment Analysis}
Our results reveal that the benefits of Transformer block skipping vary significantly across task complexity. For simple classification tasks (Tables~\ref{table:single_object_1} and \ref{table:single_object_2}), we observe that many full-layer VLMs fail to identify objects such as \textit{person}, \textit{cup}, or \textit{bird}, whereas our skipping strategy dramatically improves accuracy by removing redundant layers. This phenomenon aligns with the notion of \textit{over-thinking}, where excessive reasoning leads to output degradation in simple tasks. In contrast, for more complex tasks like image captioning (Table~\ref{tab:caption}), the performance gap between the full-layer baseline and the best skipping configuration is smaller, and the effective sparsity typically remains below 20\%. These findings suggest that block skipping is particularly powerful for lightweight tasks—especially single-object classification—where shallow layers suffice and deep layers may introduce noise or delay. GM-Skip eliminates redundant reasoning paths, leading to both improved accuracy and reduced latency in simple tasks.
\subsection{Real-World Result}
As shown in Figure~\ref{fig:real_world}, we deploy a real vehicle and simultaneously run two sets of VLMs for real-time testing. Videos of the three tasks are available in the Supplementary Material. The detailed experimental procedure is as follows:
\begin{figure}[t]
\centering
\includegraphics[width=1.0\linewidth]{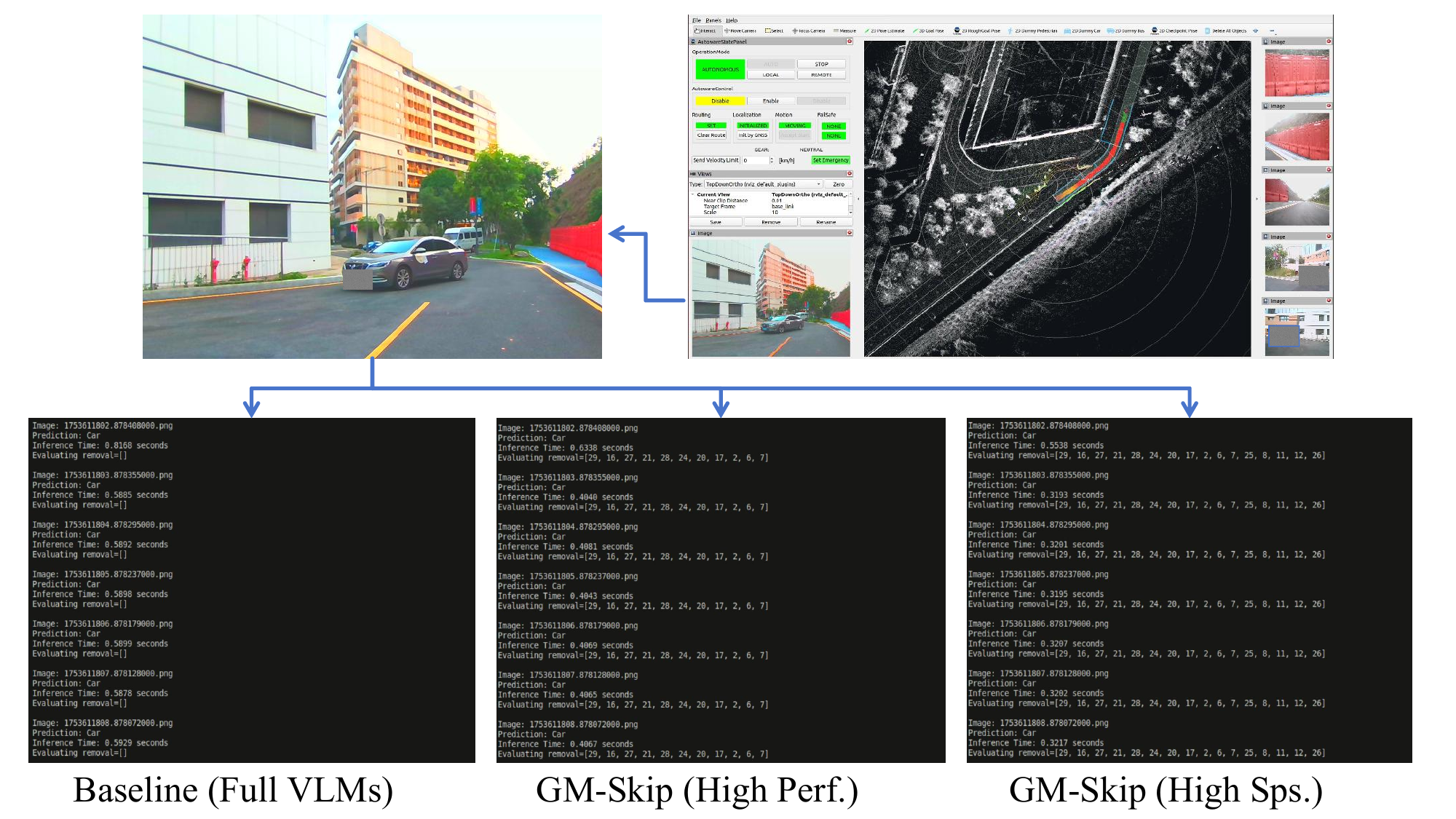}
\vspace{-15pt}
\caption{Real-world comparison between MP-Skip VLMs and Full VLMs on an autonomous vehicle. 
}
\label{fig:real_world}
\vspace{-10pt}
\end{figure}

\paragraph{Training Stage.} We train the block skipping configurations based on the CODA dataset. As shown in Table~\ref{table:real_world}, GM-Skip exhibits highly efficient skipping behavior tailored to real-world detection scenarios. In the single-object detection task, the High Sps. configuration skips \textbf{50\%} of the Transformer blocks, while still maintaining performance. In the multi-object setting, over \textbf{30\%} of the blocks are skipped. For the captioning task, although the configuration is directly transferred from COCO without retraining, it still achieves substantial sparsity and latency reduction.

\paragraph{Deployment Stage.}We deploy the trained skip configurations on our autonomous vehicle to conduct real-world tests. A front-view camera provides the image input to the VLM, and we compare the inference latency between full-layer and skip configurations under real-time conditions. As reported in Table~\ref{table:real_world}, latency is significantly reduced across all tasks: from \textbf{0.6006s} to \textbf{0.3280s} in single-object detection (\textbf{45.4\%} reduction), from \textbf{1.4265s} to \textbf{1.0170s} in multi-object detection (\textbf{28.7\%} reduction), and from \textbf{1.1293s} to \textbf{0.7106s} in captioning (\textbf{37.1\%} reduction). These real-world experiments validate the practical effectiveness of our skip configurations, showing that GM-Skip can substantially reduce inference latency without sacrificing output quality. This demonstrates its strong potential for deployment in latency-critical autonomous driving scenarios.
\section{Conclusion}
In this work, we introduced GM-Skip, a flexible and performance-aware Transformer block skipping framework tailored for Vision-Language Models. By incorporating a metric-guided greedy algorithm, reverse-order deletion, and a tunable sparsity-accuracy balance, GM-Skip enables task-specific adaptation to various VLM workloads, ranging from classification to captioning. Experimental results across COCO and CODA datasets demonstrate that GM-Skip not only surpasses existing skip strategies in both efficiency and accuracy but also remains robust in real-world autonomous driving scenarios. Our findings highlight the promise of dynamic block skipping as a practical solution for deploying VLMs in latency-sensitive applications.
\bibliography{aaai2026}


\end{document}